\documentclass[10pt,letterpaper]{article}
\usepackage[top=0.85in,left=2.75in,footskip=0.75in]{geometry}

\usepackage{amsmath,amssymb}
\usepackage{bm}
\newtheorem{problem}{Problem}

\usepackage{changepage}

\usepackage{textcomp,marvosym}

\usepackage{cite}

\usepackage{nameref,hyperref}

\usepackage[right]{lineno}

\usepackage[nopatch=eqnum]{microtype}
\DisableLigatures[f]{encoding = *, family = * }

\usepackage[table]{xcolor}

\usepackage{array}

\newcolumntype{+}{!{\vrule width 2pt}}

\newlength\savedwidth

\raggedright
\usepackage[aboveskip=1pt,labelfont=bf,labelsep=period,justification=raggedright,singlelinecheck=off]{caption}

\makeatletter
\renewcommand{\@biblabel}[1]{\quad#1.}
\makeatother

\usepackage{lastpage,fancyhdr,graphicx}
\usepackage{epstopdf}
\fancyheadoffset[L]{2.25in}
\fancyfootoffset[L]{2.25in}
\def\equationautorefname~#1\null{Eq.~(#1)\null}

\newcommand{\beginsupplement}{
    \setcounter{section}{0}
    \renewcommand{\thesection}{S\arabic{section}}
    \setcounter{equation}{0}
    \renewcommand{\theequation}{S\arabic{equation}}
    \setcounter{table}{0}
    \renewcommand{\thetable}{S\arabic{table}}
    \setcounter{figure}{0}
    \renewcommand{\thefigure}{S\arabic{figure}}
    \newcounter{SIfig}
    \renewcommand{\theSIfig}{S\arabic{SIfig}}}

\usepackage{changes}
\definechangesauthor[name={Lauri Seppäläinen}, color=blue]{LS}
\usepackage{todonotes}
\reversemarginpar
\usepackage{color}
\usepackage{booktabs}
\usepackage{doi}

\begin{document}
\vspace*{0.2in}

\begin{flushleft}
    {\Large \textbf{Using Slisemap to interpret physical data} }%
    \newline
    \\
    Lauri Sepp\"al\"ainen [0000-0002-3380-6575]\textsuperscript{\Yinyang, 1*},
    Anton Björklund [0000-0002-7749-2918]\textsuperscript{\Yinyang, 1*},
    Vitus Besel [0000-0003-4535-5422]\textsuperscript{\Yinyang, 1*},
    Kai Puolam\"aki [0000-0003-1819-1047]\textsuperscript{1*}
    \\
    \bigskip
    \textbf{1} University of Helsinki, Helsinki, Finland
    \bigskip

    \Yinyang These authors contributed equally to this work.

    * firstname.lastname@helsinki.fi
\end{flushleft}

\section*{Abstract}

Manifold visualisation techniques are commonly used to visualise high-dimensional datasets in physical sciences.
In this paper we apply a recently introduced manifold visualisation method, called {\sc slisemap}, on datasets from physics and chemistry.
    {\sc slisemap} combines manifold visualisation with explainable artificial intelligence.
Explainable artificial intelligence is used to investigate the decision processes of black box machine learning models and complex simulators.
With {\sc slisemap} we find an embedding such that data items with similar local explanations are grouped together.
Hence, {\sc slisemap} gives us an overview of the different behaviours of a black box model.
This makes {\sc slisemap} into a supervised manifold visualisation method, where the patterns in the embedding reflect a target property.
In this paper we show how {\sc slisemap} can be used and evaluated on physical data and that {\sc slisemap} is helpful in finding meaningful information on classification and regression models trained on these datasets.

\section{Introduction}\label{sec:intro}

Many real-world datasets, including those in physics, are tabular; rows correspond to data items (points) and columns to features of each data item. One problem when dealing with such datasets is making meaningful summaries. {\rm Unsupervised machine learning}, e.g., clustering or visualisation, can be used to tackle the problem of understanding datasets with many features. In manifold visualisation the objective is to find an embedding that ``compresses'' the high-dimensional data into, typically, two dimensions.
Manifold visualisation is central to exploring and understanding complex scientific datasets \cite{kobak2019,diaz2021,andronov2021Exploring,anders2018dissecting}.

Manifold visualisation methods, such as \cite{cunningham2015linear,vandenmaaten2008,2018arXivUMAP}, find embeddings such that similar data items in the original (high-dimensional) feature space end up nearby in the embedding.
A challenge when applying these visualisation methods is that the resulting embedding depends on how we measure similarities between the data points.

On the other hand, supervised learning is increasingly applied in physical sciences \cite{carleo2019mlinphysical}.
Physics simulations and powerful supervised learning modes are often ``black boxes'': it is not apparent by which criteria the model outputs are related to properties of, e.g., simulated molecules.
This paper shows examples of how quantum chemical simulations are used to estimate molecular properties and how regression models, such as random forests, can classify particle jets from high-energy physics.
Using black box models can be problematic if the aim is to understand the underlying phenomena.
Additionally, to trust the predictions, it is helpful to understand by which criteria the predictions are made.

Explainable Artificial Intelligence (XAI) is a branch of machine learning that tries to open up these black boxes; see, e.g., \cite{guidotti:2018:a,linardatos2021explainable} for reviews.
Explanation methods can be split into {\em model-specific} and {\em model-agnostic} methods.
The latter work on any supervised learning model and is what we consider here.
Explanations can further be divided into {\em global} and {\em local} explanations.
Global explanations try to explain the whole black-box model,
but here, we are interested in local explanations for individual predictions.

Many model-agnostic local explanation methods find simpler, interpretable models that locally approximate the black-box supervised learning model in the neighbourhood of the points of interest \cite{ribeiro2016,Lundberg_Lee_2017,bjorklund2023explaining}.
A common choice is to use linear models for the approximation.
Naturally, for linear models to be interpretable, the number of coefficients cannot be too large \cite{lipton2018mythos}, and the features in the data also need to be understandable \cite{wellawatte2023perspective,bjorklund2023explaining}.

Many of these methods require sampling of new data points.
If care is not taken, the sampled data points are often unrealistic and unphysical.
Evaluating black box models on data unlike the training data leads to unreliable predictions \cite{hooker2021unrestricted} and nonsensical explanations \cite{bjorklund2023explaining}.

{\sc slisemap} \cite{bjorklund2023slisemap} is a recently introduced manifold visualisation method that finds local explanations for all data items and an embedding such that nearby data items have similar explanations.
    {\sc slisemap} satisfies physical constraints since no sampling of---possibly unphysical---new data points is required.
    {\sc slisemap} produces a visualisation where items with similar explanations form visual patterns, such as clusters.

While {\sc slisemap}
is introduced in \cite{bjorklund2023slisemap}, the focus of this paper is to demonstrate its application to physical datasets.
When analysing physical data, an analyst usually has two goals: (i) to understand the structure of the high-dimensional data and (ii) how this data relates to a target property, such as a class label or saturation vapour pressure in the molecular example.
Often, a model of some kind (a simulation, a regression model, a neural network, etc.) is used to elucidate the connection between the features and the target.
As mentioned above, many model classes do not readily offer insight into how their estimate is formed.
    {\sc slisemap} offers both explanations for the target property in the form of local simple models while also providing a visualisation where data points with similar explanations form patterns.

The objectives of this paper are the demonstration of (i) how {\sc slisemap} can be applied to describe physical datasets (\autoref{sec:workflow}), (ii) how the resulting visualisations can be evaluated (\autoref{sec:workflow}), and (iii) that the explanations carry meaningful information for the domain expert (\autoref{sec:usage}).

\section{Related work}\label{sec:related}
{\sc slisemap} \cite{bjorklund2023slisemap} is a method for manifold visualisation and explainable artificial intelligence (XAI).
This section briefly reviews related work on these topics and how they are used in science.

Manifold visualisation is commonly applied to explore and understand complex data in many fields of science, from genetics \cite{kobak2019,diaz2021} and chemoinformatics \cite{andronov2021Exploring} to astronomy \cite{anders2018dissecting} and linguistics \cite{levine2020sensebert}.
Generally, manifold visualisation is used to find structures, such as clusters, in large and high-dimensional datasets.

Manifold visualisation aims to present high-dimensional data as a low-dimensional (usually 2D) embedding while preserving the maximum amount of information based on preset criteria.
In science, standard manifold visualisation methods include linear projections \cite{cunningham2015linear}, such as principal component analysis (PCA) \cite{pearson1901liii}, and non-linear methods, such as t-SNE \cite{vandenmaaten2008}, and UMAP \cite{2018arXivUMAP}.
These are examples of unsupervised methods; they can be utilized to discover structure in data but cannot offer insight into how the features affect the predictions.
For a survey on manifold visualisation techniques, see, e.g., \cite{sorzano2014survey}.
There is also a supervised variant of UMAP \cite{2018arXivUMAP}, and several other supervised manifold visualisation techniques have been proposed, such as linear low-rank projections (LOL) \cite{vogelstein2021Supervised}.

Results from manifold visualisation methods are often analysed via visual inspection.
Hence, we need to verify that the emerging patterns reflect some true quality of the data and are not just artefacts produced by the algorithm in question.
A wide range of visualisation evaluation metrics has been proposed, from summary statistics such as stress \cite{joia2011affine} to visualisations such as Shepard diagrams \cite{leeuw2015shepard}.
However, many evaluation metrics rely on the notion of distance both in the input data and the visualisation.
In science, data is often tabular, and features can combine class variables, vectors and scalar values with different units.
Defining suitable distance measures is, therefore, challenging.
Furthermore, not all manifold visualisation methods produce linear or even metric embeddings.

Evaluation metrics inspired by information retrieval have been proposed for non-linear embedding methods, such as precision and recall \cite{JMLR:v11:venna10a}.
Incidentally, the definition of precision, or how neighbouring points in the input space are neighbours in the embedding space, serves as a basis for the neighbourhood quality metric we introduce in \autoref{sec:workflow}.
For a thorough quantitative survey concerning the evaluation of visualisations, see \cite{espadoto2021Quantitative}.

Explainable AI most often seeks to enhance the interpretability of black-box models.
XAI methods can be classified along two axes: \emph{model-specific} vs. \emph{model-agnostic} and \emph{local} vs. \emph{global}.
Model-specific methods can only be applied to single (or class of) models, while model-agnostic methods can be used for any model.
Global methods aim to summarise the entire black-box model, while local methods explain limited regions of the data manifold, e.g., near a selected data item.

As black-box models can be very complex, summarising them holistically and interpretably is challenging.
As such, many model-agnostic explanation methods provide only local explanations.
A common approach is to use a interpretable surrogate model to locally approximate the black-box model \cite{guidotti:2018:a}.

These methods include {\sc lime} \cite{ribeiro2016}, {\sc shap} \cite{Lundberg_Lee_2017} and {\sc slise} \cite{bjorklund2023explaining}.
{\sc lime} provides sparse linear models as explanations for classifiers, with a penalty function for local model complexity to ensure interpretability.
The method learns these sparse local models by sampling (generating) new data points around existing ones and utilizes the complex model to predict the target property for these new points.
    {\sc shap} uses additive feature attributions (linear models on binary variables) where the feature contributions are estimated using ideas from cooperative games in game theory.
Like {\sc lime}, {\sc shap} also relies on sampling new data items near the input points.
Sampling-based methods, however, have a weakness when analysing datasets with underlying constraints.
Such constraints can incorporate, e.g., physical conservation laws such as energy conservation.

    {\sc slise} \cite{bjorklund2019,bjorklund2023explaining} is an XAI method explicitly designed to not rely on sampling new data.
The method is based on robust regression.
    {\sc slise} splits the data into a set of potential outliers and a set of non-outliers.
This translates to explanations if we consider items with different local explanations as outliers (with respect to a selected data item).
Non-outlier points then specify a linear model, while outlier points are ignored.

For a broader review of XAI methods, we refer to, e.g., \cite{linardatos2021explainable,guidotti:2018:a} for surveys of XAI methods and \cite{wellawatte2023perspective} for XAI in chemistry.

\section{Methods}\label{sec:methods}

In this section, we introduce the methods for this paper.
We begin, in \autoref{sec:slisemap}, with a brief overview of {\sc slisemap} \cite{bjorklund2023slisemap}. and in \autoref{sec:workflow}, we describe how we use {\sc slisemap} and define several evaluation criteria.

\subsection{SLISEMAP}\label{sec:slisemap}

{\sc slisemap} \cite{bjorklund2023slisemap} is a tool for visualising all local explanations in a dataset.
Specifically, explanations are interpretable models that locally approximate the complex model, similar to \cite{bjorklund2023explaining,ribeiro2016,Lundberg_Lee_2017}.
{\sc slisemap} finds low-dimensional embedding and local models (explanations) such that items with similar local models end up next to each other.
In {\sc slisemap}, the local models are not limited to any single class; however, in this paper, we only consider simple linear regression and classification models.
Formally, {\sc slisemap} solves the following problem:
\begin{problem}
\label{prob:slisemap}
Assume you are given a dataset of $n$ data items $(\bm x_1, \bm y_1),\ldots,(\bm x_n, \bm y_n)$, where $\bm x_i \in \mathbb{R}^m$ are the vectors of features and $\bm y_i \in \mathbb{R}^o$ are the targets,
and a radius $r\in{\mathbb{R}}_{>0}$.
For every data item $i \in \{1, \ldots, n\}$, find the embedding $\bm z_i \in \mathbb{R}^d$ and local model $f_i : \mathbb{R}^m \rightarrow \mathbb{R}^o$ that minimise the loss
\begin{multline}
    \label{eq:slisemap}
    \mathcal{L} = \sum\nolimits_{i = 1}^{n} \sum\nolimits_{j = 1}^{n}
    \frac{ e^{-D(\bm z_i, \bm z_j)} }{ \sum_{k=1}^{n} e^{-D(\bm z_i, \bm z_k)} }
    l(f_i(\bm x_j), \bm y_j) \\ + \sum\nolimits_{i=1}^n \sum\nolimits_{j=1}^{p} (\lambda_{\textrm{lasso}} |\mathbf{B}_{ij}| + \lambda_{\textrm{ridge}} \mathbf{B}_{ij}^2),
\end{multline}
where $D(\cdot)$ is the Euclidean distance and $l(\cdot,\cdot)$ is a loss function for the local models under the constraint that
\begin{equation}
    \label{eq:radius}
    \sum\nolimits_{i=1}^{n}\sum\nolimits_{k=1}^{d}{\bm z}_{ik}^2 / n = r^2.
\end{equation}
The rows of $\mathbf{B} \in \mathbb{R}^{n \times p}$ contain the parameters for the local models $f_i$, where $p$ is the number of parameters in the local white-box models, and $\lambda_{\textrm{lasso}} \geq 0$ and $\lambda_{\textrm{ridge}} \geq 0$ are the parameters for Lasso and Ridge regularisation, respectively.
\end{problem}

Here we consider 2-dimensional embeddings ($d=2$) with radius $r=3.5$, as recommended by \cite{bjorklund2023slisemap}.
We use linear regression for $f_i$ and mean squared error as $l$ for regression problems.
With classification problems, we use logistic regression for $f_i$ and Hellinger loss as $l$; see \cite{bjorklund2023slisemap} for details. We optimise \autoref{eq:slisemap} using \textsc{lbfgs} \cite{nocedal1980updating} combined with a greedy heuristic for escaping local optima by using the {\sc slisemap} library \cite{bjorklund2023slisemapdemo}.\footnote{\url{https://github.com/edahelsinki/slisemap}} We have used $\chi$iplot visualisation platform for data exploration \cite{tanaka2023chi}.

\subsection{Workflow and Performance Measures}
\label{sec:workflow}

We start by normalising the data into zero mean and unit variance.
This lets us use the built-in $L^1$ and $L^2$ regularisations for the local models to avoid overfitting.
$L^1$ regularisation typically yields sparse solutions \cite{roberttibshiraniRegressionShrinkageSelection1996}, making interpretation easier \cite{guidotti:2018:a}.

{\sc slisemap} produces an embedding and a local explanation for each data item. To verify that the embedding and local models are reliable, we introduce three performance measures to test for various aspects of the solution: (i) permutation loss (a sanity check), (ii) local model stability (whether the found local models are stable), and (iii) neighbourhood stability (whether points have a stable and unique local model).
We use these measures in \autoref{sec:stability}.

\subsubsection*{Permutation Loss}
The {\em permutation loss} measures whether the {\sc slisemap} solution captures a connection between the features and the target variable.
More specifically, denote by ${\cal L}_{permuted}$ the loss of \autoref{eq:slisemap} on a {\sc slisemap} solution optimised after the target variable ${\bf y}$ has been randomly permuted, and by ${\cal L}$ the loss on the original unpermuted data.
We define {\em permutation loss} as
\begin{equation}
    {\cal M}_{permutation}={\cal L}/{\cal L}_{permuted}.
\end{equation}
We expect ${\cal M}_{permutation}<1$; otherwise, the {\sc slisemap} has not been able to capture any relations between the features and the target variable. The permutation loss acts as a sanity check for the solution.

\subsubsection*{Local Model Stability} The {\em local model stability} measures the stability of the found local models with respect to the resampling of the data, i.e., can we trust the local models?
Assume we have obtained two datasets of the same size, e.g., by resampling $n$ data points without replacement from the original data, resulting in two sets of local models $\{f_1,\ldots, f_n\}$ and $\{f'_1,\ldots,f'_n\}$.
To compare the similarity of the local model populations, we use the Hungarian algorithm to find a permutation $\pi$ such that the sum of distances is minimised:
\begin{equation}
    {\cal M}_{models}=\min_\pi \sum\nolimits_{i=1}^n{D(f_i,f'_{\pi(i)})} ~/ Z,
\end{equation}
where $D(f_i,f'_{\pi(i)}) = ||\mathbf{B}_{i, \cdot} - \mathbf{B}_{\pi(i), \cdot}||$ is the Euclidean distance between the coefficients of the local models and $Z=\sum\nolimits_{i=1}^n{\sum\nolimits_{j=1}^n{D(f_i, f'_j) / n}}$.

\subsubsection*{Neighbourhood Stability}
The {\em neighbourhood stability} measures the stability of the neighbourhoods with respect to the resampling of the data, i.e., is the relative location of individual data points stable under resampling?
To compute neighbourhood stability, we sample two datasets of size $n$, which share 50~\% of the data items.
Denote by ${\cal S}$ the set of points shared by these two datasets, and by $\{\bm z_1,\ldots,\bm z_n\}$ and $\{\bm z'_1,\ldots,\bm z'_n\}$ the points in the two embeddings.
We define the neighbourhoods of point $i$ in the first embedding by $N(i)=\{j\in{\cal S}\mid D(\bm z_i,\bm z_j)\le 1\}$ and $N'(i)=\{j\in{\cal S}\mid D(\bm z'_i,\bm z'_j)\le 1\}$ in the second embedding.
The {\em neighbourhood stability} is defined as an average Jaccard similarity between the neighbourhoods:
\begin{equation}
    {\cal M}_{neighbourhood}=1 - {\left|{\cal S}\right|}^{-1}
    \sum\nolimits_{i\in{\cal S}}{
    {\left|N(i)\cap N'(i)\right|}/{\left|N(i)\cup N'(i)\right|}
    }
\end{equation}

\subsubsection*{Explanation quality}
In addition to the metrics described above, we want to measure the quality of the explanations produced by {\sc slisemap}.
To measure explanation quality, we use three metrics: local loss, local loss with nearest neighbours and coverage with nearest neighbours.
These metrics are described in \cite{bjorklund2023slisemap}, and we have included their definitions in the Supporting Information (\autoref{sec:explanation_metrics}).
Some metrics require choosing additional parameters; in such cases, we use values found in \cite{bjorklund2023explaining}.

Local loss measures how well the local models fit their targets (or, in case of nearest neighbours local loss, those of their neighbours).
Coverage measures how well the local models generalise to their neighbours.
For comparison, we also calculate these values for other widely used manifold visualisation methods, namely PCA, t-SNE and UMAP.
To produce explanations for these alternative methods, we first calculate an embedding with each one and then train local models (similar to {\sc slisemap}) based on that embedding.
The results from the comparison can be found in \autoref{tbl:DR_comp}.

\subsubsection*{Analysis}
{\sc slisemap} is foremost designed as a tool for investigating and understanding black box models.
Therefore, the main way to analyse the solutions is through plots \cite{tanaka2023chi}.
We demonstrate how to analyse {\sc slisemap} solutions for multiple datasets in \autoref{sec:usage} and how domain expertise can help interpret and validate the solutions.

Analysing potentially thousands of local models is not feasible.
Hence, we often cluster the local models using {\sc k-means} clustering in the local model parameter space, i.e., the rows of the matrix $\mathbf{B}$.
If a suitable number of clusters is chosen, the cluster centroids are a good proxy for most local models.
However, data items in lower-density areas of the embedding might have very different local models from the cluster centroids.
Since they also have small neighbourhoods, the local models are at risk of overfitting, and care should be taken when interpreting these points. Notice that we use {\sc k-means} clustering after {\sc slisemap} only to help visualise and interpret the results.

\section{Use Cases} \label{sec:usage}

This section demonstrates how {\sc slisemap} can be applied to physical datasets and how we analyse and verify the solutions using domain knowledge and the metrics from above.\footnote{The datasets and source code used in the paper are available at \url{https://www.edahelsinki.fi/papers/slisemap\_phys}.}
In \autoref{sec:gecko}, we study a dataset from atmospheric science; in \autoref{sec:jets} a dataset from high energy physics, and in \autoref{sec:qm9} a dataset about organic chemistry.
Finally, in \autoref{sec:stability}, we evaluate the solutions using the performance measures from \autoref{sec:workflow}.

\subsection{Atmospheric relevant organic molecules: GeckoQ} \label{sec:gecko}
We applied {\sc slisemap} to 31,637 atmospheric relevant organic compounds contained in the \textit{GeckoQ} dataset \cite{besel2023GeckoQ}, collected by one of the authors.
Each molecule has a variety of properties, such as the number of specific functional groups, \textit{saturation vapour pressure} ($p_\mathrm{Sat}$) and \textit{topological fingerprint} (TopFP) descriptor \cite{Nilakantan87,rdkit}.

The $p_\mathrm{Sat}$ is a measure of the affinity of a molecule to condense into the liquid phase or to remain in the gas phase. This property is relevant in atmospheric science because low-volatile organic compounds are known to be driving factors for particle formation in the atmosphere \cite{kerminen2018}. The GeckoQ $p_\mathrm{Sat}$ have been calculated with the \textit{Conductor-like Screening Model for real solvents} (COSMO-RS)\cite{Klamt93,Klamt98}, a quantum chemistry-based method, and will be the (log-transformed) target variables in following {\sc slisemap} embeddings.
A subset of molecular properties was chosen as \textit{interpretable features} to construct the {\sc slisemap} embeddings. A list of all explainable features can be found in the Supporting Information (\autoref{sec:exp_feats}).

\begin{figure}[!tb]
    \centering
    \includegraphics[width=\textwidth]{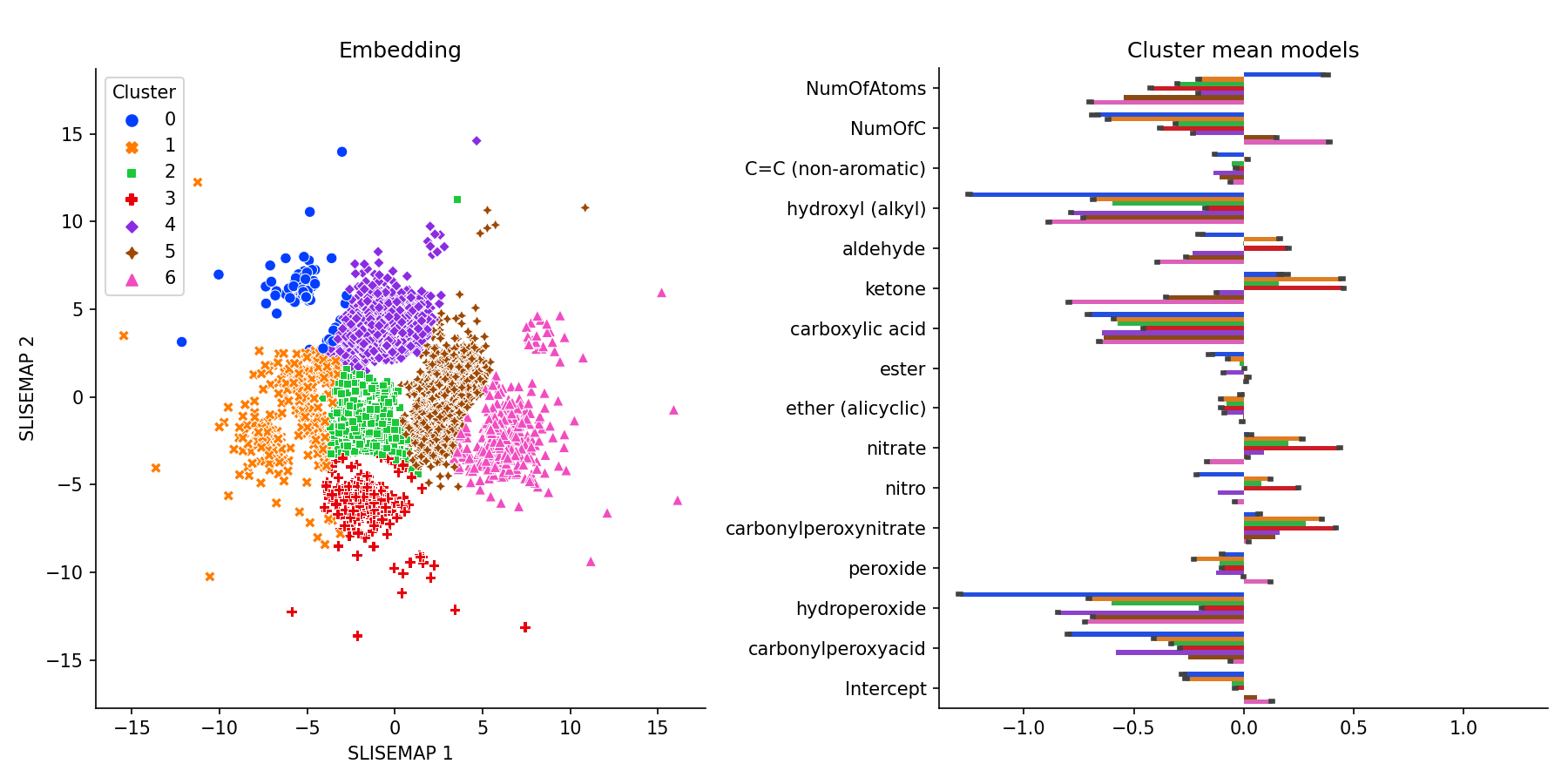}
    \caption{The {\sc slisemap} embedding of the GeckoQ data in the left panel. The number of clusters (seven) was chosen via visual inspection. The right panel includes the average local coefficients of each cluster.}
    \label{fig:SM_ExpF_real_clusters}
\end{figure}

\autoref{fig:SM_ExpF_real_clusters} shows a {\sc slisemap} embedding of the GeckoQ data. The local model coefficients generally match chemical expectations, meaning \textit{functional groups} (FGs) that are especially known to lead to a low $p_\mathrm{Sat}$ have the most negative coefficients in their local models: hydroxyl, hydroperoxide, carboxylic acid, and carbonylperoxyacids. These FGs can form hydrogen bonds, the strongest type of inter-molecular dipole-dipole interactions, and thus, the molecules are particularly strongly bound within the liquid phase, which leads to a low $p_\mathrm{Sat}$.
The incidence of each FG by cluster can be found in the Supporting Information (\autoref{fig:incidence_by_cluster}).

\begin{figure}[!tb]
    \centering
    \includegraphics[width=\textwidth]{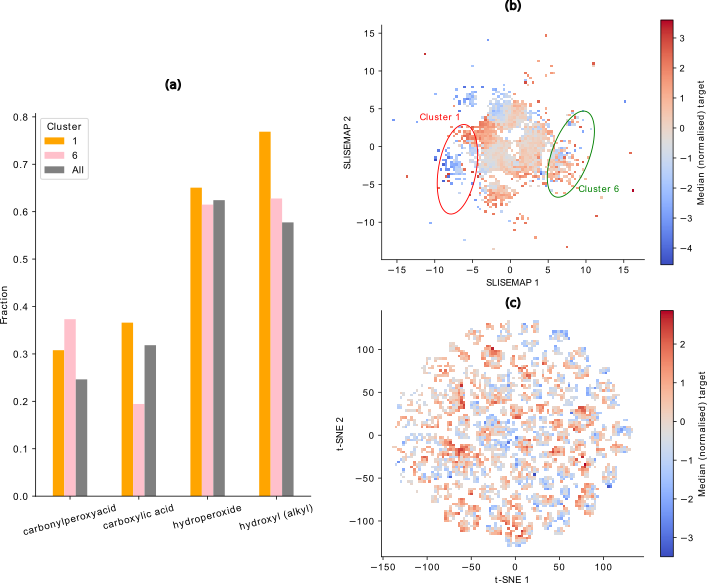}
    \caption{(a) Fraction of molecules that contain at least one FG of hydroxyl, hydroperoxide, carboxylic acid, carbonylperoxyacid grouped by clusters 1, 6 and all the clusters. (b) {\sc slisemap} and (c) t-SNE embedding, where the data points are binned, and the colour map corresponds to the median normalised target of the bins. Clusters 1 and 6 are encircled.}
    \label{fig:FG}
\end{figure}

To further substantiate the {\sc slisemap} embedding's agreement with the chemical expectancies, we chose cluster 1 (orange) and cluster 6 (pink), two visually distinct clusters for detailed analysis.
In cluster 1, the median target (non-normalised) is $2.15 \cdot 10^{-9}$~mbar (1686 molecules), whereas cluster 6 contains molecules with a higher median $p_\mathrm{Sat}$, $5.75 \cdot 10^{-6}$~mbar (1862 molecules).
The median $p_\mathrm{Sat}$ for all the data is $1.55 \cdot 10^{-6}$~mbar.
\autoref{fig:FG}a depicts the fraction of molecules in the chosen clusters and the overall data containing FGs forming hydrogen bonds.
Cluster 1 contains considerably more carboxylic acid groups than cluster 6 and the overall data, which can be directly linked to its particularly low median $p_\mathrm{Sat}$.
Additionally, cluster 1 shows a higher incidence of hydroxyl groups, while cluster 6 has a higher fraction of carbonylperoxyacid groups.
Both clusters have a similar fraction of hydroperoxide groups.

Differences between clusters also become apparent through binning the data points in the {\sc slisemap} embedding and analysis of the median (normalised) target values in the bins (cf. \autoref{fig:FG}b). Generally, regions of low and high $p_\mathrm{Sat}$ emerge. These regions roughly correspond to the clusters determined by {\sc slisemap} in \autoref{fig:SM_ExpF_real_clusters}, which is the most distinct in the points corresponding to clusters 1 and 6. The existence of these regions means that {\sc slisemap} can capture this target-feature relationship.
For comparison, \autoref{fig:FG}c depicts the binned logarithm of the median target for a t-SNE embedding, an alternative way of visualising data projected into a two-dimensional space that is popular in the field of chemoinformatics.
t-SNE embeds the data compactly with a tail in the negative and positive x-direction.
Unlike the {\sc slisemap} embedding, the t-SNE does not display distinct regions of low and high $p_\mathrm{Sat}$.
This is further reflected in the ranges of the median values: t-SNE has a smaller range ($-3.49$ to $2.86$) than {\sc slisemap} ($-4.56$ to $3.60$).

Furthermore, the explanation quality metrics (see \autoref{tbl:DR_comp}) show that {\sc slisemap} scores higher in all three explanation quality metrics than t-SNE.
This reinforces how {\sc slisemap} is able to prioritise patterns related to the target when building the embedding, whereas t-SNE, due to its unsupervised nature, cannot group molecules with similar behaviour together.

\subsection{Elementary Particle Jets} \label{sec:jets}

This dataset contains simulated LHC proton-proton collisions \cite{cms2017dataset}.
The collisions create elementary particles such as \emph{quark}s and \emph{gluon}s.
Quarks and gluons decay into cascades of stable particles, called jets, before being detected.
The classification task is to determine if the particle that created a jet was a quark or a gluon \cite{cms2013performance}.

To get probabilities for the jets, we train a random forest classifier \cite{breiman2001random} with $100$ trees and $50$ leaves per tree.
When applying {\sc slisemap} \cite{bjorklund2023slisemap} on this dataset, we use logistic regression as the local models $f_i$ in \autoref{eq:slisemap}.

\begin{figure}[htbp]
    \centering
    \includegraphics[width=\textwidth]{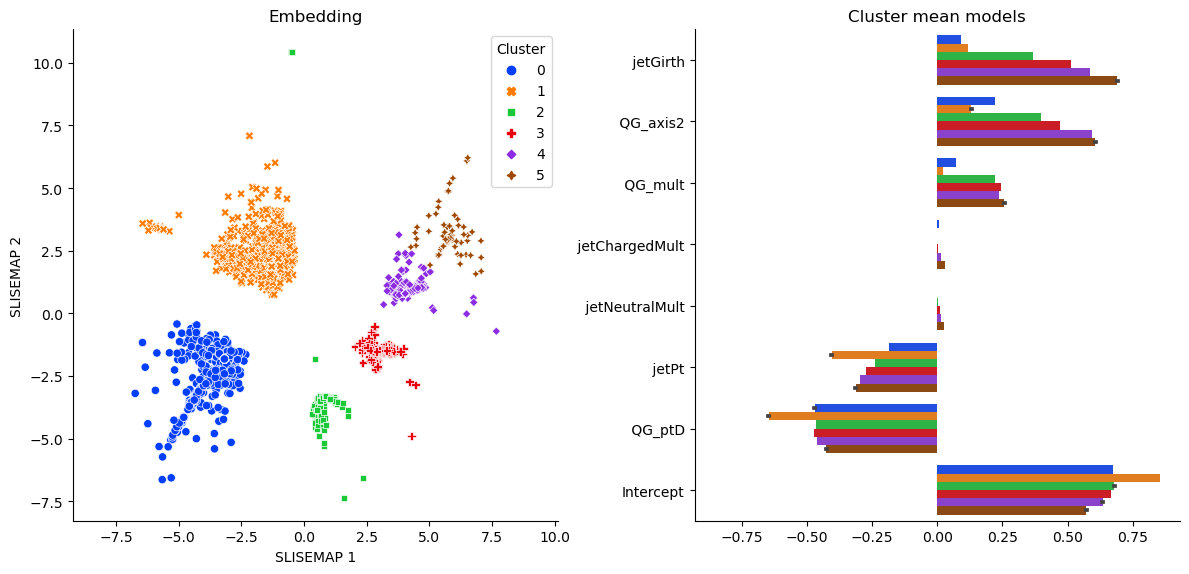}
    \caption{
        {\sc slisemap} solution for $10~000$ jets from the particle jets dataset.
        The data items have been clustered according to the local models.
        The coefficients for all local models follow physical theory but with varying magnitudes.
    }
    \label{fig:jets}
\end{figure}

The resulting solution can be seen in \autoref{fig:jets}.
The local models match the underlying quantum chromodynamics \cite{cms2013performance}.
Wider jets (high \texttt{jetGirth} and \texttt{QG\_axis2}) and jets with more particles (\texttt{QG\_mult}) are more gluon-like.
Splitting the multiplicity based on charge should not offer any more information.
The total momentum (\texttt{jetPt}) usually does not matter, except for higher energies where quarks are more likely.
How much of the total momentum parallels the jet (\texttt{QG\_ptD}) is also helpful in finding quarks.

Even though all local models match the theory, there are some differences in magnitudes.
E.g., the blue cluster 0, in \autoref{fig:jets}, contains most of the highest energy jets.
Since multiplicity and, indirectly, the spread is also dependent on the energy; these variables are less helpful in this subset.
In contrast, the orange cluster 1 contains many jets easily classified as gluons based on \texttt{QG\_ptD} alone.

Summarising, the SLISEMAP explanations are consistent with the physical knowledge (gluon jets are generally wider) and provided additional insights into the features used by the random forest classifier.

\subsection{Small Organic Molecules: QM9} \label{sec:qm9}

\begin{figure}[htbp]
    \centering
    \includegraphics[width=\textwidth]{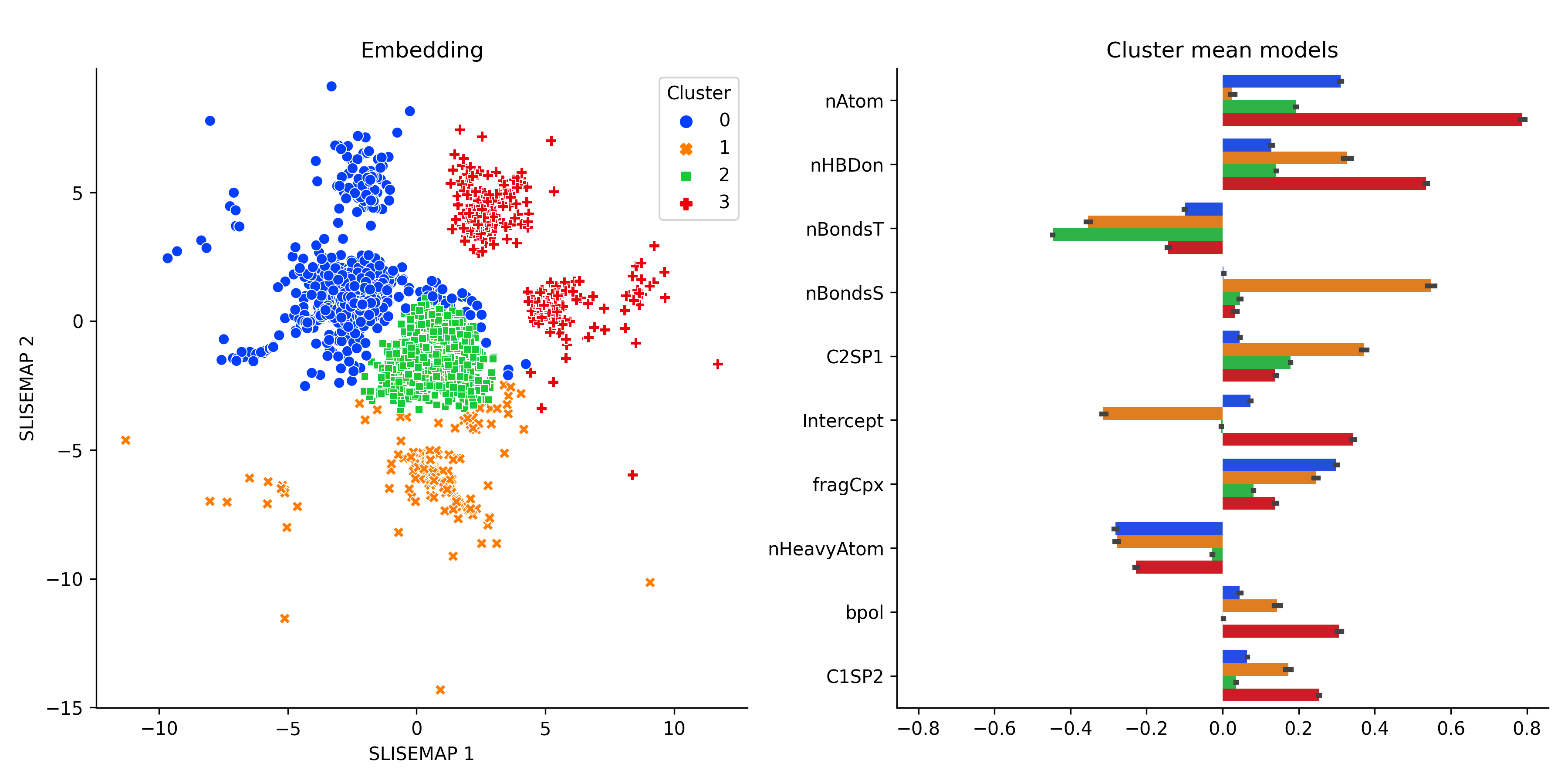}
    \caption{{\sc slisemap} embedding for the QM9 data set (10,000 molecules), clustered with 4 clusters. The right panel shows the ten most influential features for predicting the target (HOMO energy).}
    \label{fig:qm9}
\end{figure}

Finally, we applied {\sc slisemap} to the molecules in the QM9-dataset\cite{blum2009,rupp2012}, a data set of 133,814 small organic molecules.
We utilised HOMO energies obtained from \cite{stuke2019chemical,ghosh2020qm9} as targets and created interpretable features with the Mordred molecular descriptor calculator\cite{Moriwaki2018}.
Here, we will focus on qualitative analysis based on the {\sc slisemap} embedding and omit a more technical chemical analysis.
\autoref{fig:qm9} displays a similar embedding structure to the GeckoQ embedding: the bulk of the data points group around a few centres, and a few data points are spread out.
However, unlike GeckoQ, the clusters are much more distinct.
Setting the number of clusters to four, we can see a central cluster flanked by three others.
For the central cluster, the number of triple bonds (\texttt{nBondsT}) has relatively higher importance than for the other clusters, as can be seen in the right panel in \autoref{fig:qm9}.

The orange and red clusters place high importance on the number of hydrogen bond donors (\texttt{nHBDon}).
They are distinguished from one another by the number of atoms (\texttt{nAtoms}), which is of high importance to the red cluster while nearly inconsequential for the orange one, and the number of sulphur bonds (\texttt{nBondsS}), which sets the orange cluster apart from the others.

\subsection{Evaluation of the solutions}\label{sec:stability}
\begin{table}
    \caption{Comparison of explanation measures for {\sc slisemap}, PCA, t-SNE and UMAP for the datasets described in this paper; computed as described in \cite{bjorklund2023slisemap}. Bold values indicate the best performance, and the error bounds the standard deviation with respect to the resampling of the data.}
    \label{tbl:DR_comp}
    \centering
    \begin{tabular}{llll}
        \toprule
        Model          & Local loss $\downarrow$               & NN Local loss $\downarrow$            & NN Coverage $\uparrow$   \\
        \midrule
        Dataset:    GeckoQ                                                                                                        \\
        \midrule
        {\sc slisemap} & $\mathbf{0.03 \pm 0.01}$              & $\mathbf{0.04 \pm 0.01}$              & $\mathbf{0.79 \pm 0.06}$ \\
        PCA            & $0.28 \pm 0.01$                       & $0.29 \pm 0.00$                       & $0.31 \pm 0.00$          \\
        t-SNE          & $0.28 \pm 0.00$                       & $0.30 \pm 0.01$                       & $0.31 \pm 0.00$          \\
        UMAP           & $0.29 \pm 0.01$                       & $0.30 \pm 0.01$                       & $0.30 \pm 0.00$          \\
        \midrule
        Dataset:    Jets                                                                                                          \\
        \midrule
        {\sc slisemap} & $\mathbf{1.8 \cdot 10^{-4} \pm 0.00}$ & $\mathbf{2.9 \cdot 10^{-4} \pm 0.00}$ & $\mathbf{0.87 \pm 0.02}$ \\
        PCA            & $6.2 \cdot 10^{-4} \pm 0.00$          & $1.0 \cdot 10^{-3} \pm 0.00$          & $0.57 \pm 0.00$          \\
        t-SNE          & $6.8 \cdot 10^{-4} \pm 0.00$          & $1.2 \cdot 10^{-3} \pm 0.00$          & $0.56 \pm 0.00$          \\
        UMAP           & $1.3 \cdot 10^{-3} \pm 0.00$          & $1.5 \cdot 10^{-3} \pm 0.00$          & $0.43 \pm 0.00$          \\
        \midrule
        Dataset:    QM9                                                                                                           \\
        \midrule
        {\sc slisemap} & $\mathbf{0.03 \pm 0.01}$              & $\mathbf{0.06 \pm 0.01}$              & $\mathbf{0.74 \pm 0.05}$ \\
        PCA            & $0.22 \pm 0.00$                       & $0.26 \pm 0.00$                       & $0.36 \pm 0.00$          \\
        t-SNE          & $0.22 \pm 0.00$                       & $0.26 \pm 0.00$                       & $0.35 \pm 0.00$          \\
        UMAP           & $0.23 \pm 0.00$                       & $0.25 \pm 0.00$                       & $0.37 \pm 0.00$          \\
        \bottomrule
    \end{tabular}
\end{table}

\begin{figure}[tbp]
    \centering
    \includegraphics[width=\textwidth]{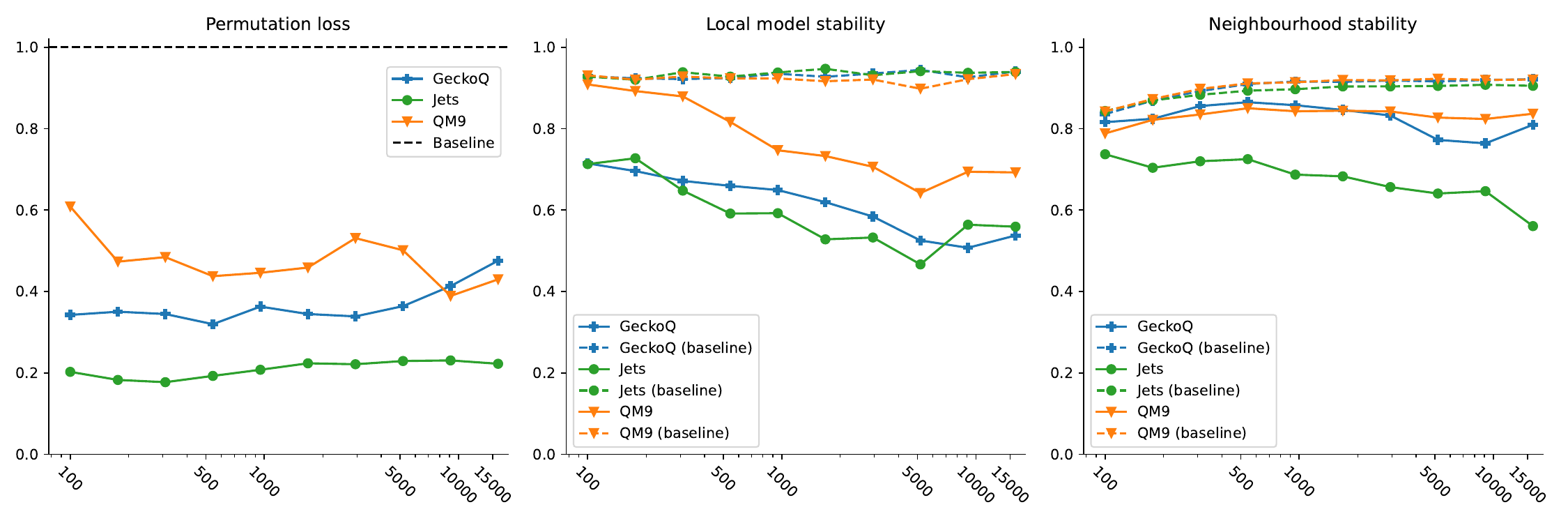}
    \caption{Permutation loss, local model stability and neighbourhood stability (\autoref{sec:workflow}) for the use cases as a function of (resampled) dataset size.
        Smaller values are better.
        Dashed lines indicate reference values for baseline datasets where the target variables have been permuted randomly.
    }
    \label{fig:stability_measures}
\end{figure}

It is essential to evaluate whether the resulting visualisations contain useful information or whether we are just visualising random noise. The three performance measures described in \autoref{sec:workflow} are designed to evaluate various aspects of visualisation. We depict the performance measures as a function of the sample size in \autoref{fig:stability_measures}.
For each measure and subsampled size, we train ten models and average over them to get the final performance measure value.
To provide a baseline for the stability measures, we also calculate the measure between {\sc slisemap}s trained on actual data and those trained on data with randomly permuted target variables.

The permutation loss for each dataset remains less than one (the permuted baseline), assuring that {\sc slisemap} has learned from the data.
The local model stability shows similar behaviour, rapidly falling below the baseline as sample size increases, especially for the particle jets dataset.
The measure also suggests that the sample sizes (10,000, 10,000 and 31,637 for the QM9, jets and Gecko datasets, respectively) are reasonable.

The neighbourhood stability is reasonable for the particle jets data set but worse for the molecular datasets.
But, as noted in \cite{bjorklund2023slisemap}, a data item might have multiple (almost) equally viable explanations.
We argue that the presence of these alternative explanations is inherent to local explanation methods, including {\sc slisemap}.
When we use linear models as {\sc slisemap} explanations, data items near the intersection of the local model hyperplanes could be explained almost equally well by both intersecting models.
As a practical consequence, one should be careful when, e.g., studying individual molecules in the {\sc slisemap} clusters (as the molecule could potentially also  be allocated to another cluster).
However, even for the molecular dataset, the set of local models is stable.

It follows that all {\sc slisemap} embeddings are non-random (permutation loss is less than one), and the set of local models (local model stability) is informative. However, for both molecular datasets, a single molecule can have several local explanations (neighbourhood stability), a known feature of local explanations, which SLISEMAP makes visible (see \cite{bjorklund2023slisemap}, Sect. 4.5, for further discussion).

As for explanation quality, \autoref{tbl:DR_comp} shows that {\sc slisemap} produces best explanations for each of the three datasets based on all of the chosen quality metrics.

\section{Discussion and Conclusions}\label{sec:discussion}

In this paper, we apply {\sc slisemap} \cite{bjorklund2023slisemap}, a recent XAI and visualisation method, on three scientific datasets.
Using domain knowledge, we verify that the explanations are consistent with the chemical and physical theories.

    {\sc slisemap} captures the chemical expectations set for the GeckoQ data. The local models match the different natures of the functional groups. Further, {\sc slisemap} can find underlying structures in the atmospheric data and reflect the relationship between the target and the features, which are generally known but hardly quantifiable.
The local models in the Jets dataset also adhere to (known) physics. However, the subgroups with slightly different models were new but, after analysis, justifiable.
For the QM9 dataset, structure emerges in the {\sc slisemap} embedding with distinctively behaving clusters of molecules with different local explanations for each of the clusters.

In \autoref{sec:workflow}, we introduce stability measures for the {\sc slisemap} embedding.
Our stability measures reveal that the explanations for individual data items are often not unique, a fact that {\sc slisemap} (unlike most other local explanation methods) makes apparent \cite{bjorklund2023slisemap}.
However, in \autoref{sec:stability}, we show that the collection of explanations (local models) found by the {\sc slisemap} are informative and stable for all our use cases.
Moreover, \autoref{tbl:DR_comp} shows that {\sc slisemap} captures patterns for predicting the target value better than other manifold visualisation methods widely used in these domains, as measured through the local models.

As mentioned in \autoref{sec:stability}, there is ambiguity in the selection of local explanations.
Nonetheless, it would be interesting to, in the future, analyse the alternative local explanations in depth.
Furhtermore, our results suggest that while explanations for individual points are not guaranteed stability, the overall distribution of local models should be.
Constructing these sets of plausible local models and, by extension, approximating the overall distribution could allow for uncertainty quantification.

In conclusion, we have demonstrated a workflow for applying {\sc slisemap} and shown how {\sc slisemap} can provide physically sound insight for analysing complex physical datasets.

\subsubsection*{Acknowledgements}

We thank Dr Jarmo M\"akel\"a, Prof. Patrick Rinke, and Dr Hilda Sandstr\"om for helpful discussions.
We thank the Research Council of Finland (decisions 346376, 345704, 337549, and 346368) and the Doctoral Programme in Computer Science at University of Helsinki for funding, the University of Helsinki library for support with the publication fees, and the Finnish Computing Competence Infrastructure (FCCI) for supporting this project with computational resources.

\bibliography{ms}

\begin{thebibliography}{10}

\bibitem{kobak2019}
Kobak D, Berens P.
\newblock The Art of Using T-{{SNE}} for Single-Cell Transcriptomics.
\newblock Nature Communications. 2019;10(1):5416.
\newblock doi:{10.1038/s41467-019-13056-x}.

\bibitem{diaz2021}
{Diaz-Papkovich} A, {Anderson-Trocm{\'e}} L, Gravel S.
\newblock A Review of {{UMAP}} in Population Genetics.
\newblock Journal of Human Genetics. 2021;66(1):85--91.
\newblock doi:{10.1038/s10038-020-00851-4}.

\bibitem{andronov2021Exploring}
Andronov M, Fedorov MV, Sosnin S.
\newblock Exploring {{Chemical Reaction Space}} with {{Reaction Difference
  Fingerprints}} and {{Parametric}} T-{{SNE}}.
\newblock ACS Omega. 2021;6(45):30743--30751.
\newblock doi:{10.1021/acsomega.1c04778}.

\bibitem{anders2018dissecting}
Anders F, Chiappini C, Santiago BX, Matijevi{\v c} G, Queiroz AB, Steinmetz M,
  et~al.
\newblock Dissecting Stellar Chemical Abundance Space with T-{{SNE}}.
\newblock Astronomy \& Astrophysics. 2018;619:A125.
\newblock doi:{10.1051/0004-6361/201833099}.

\bibitem{cunningham2015linear}
Cunningham JP, Ghahramani Z.
\newblock Linear {{Dimensionality Reduction}}: {{Survey}}, {{Insights}}, and
  {{Generalizations}}.
\newblock Journal of Machine Learning Research. 2015;16(89):2859--2900.

\bibitem{vandenmaaten2008}
van~der Maaten L, Hinton G.
\newblock Visualizing {{Data}} Using T-{{SNE}}.
\newblock Journal of Machine Learning Research. 2008;9(86):2579--2605.

\bibitem{2018arXivUMAP}
McInnes L, Healy J, Melville J. {{UMAP}}: {{Uniform Manifold Approximation}}
  and {{Projection}} for {{Dimension Reduction}}; 2020.

\bibitem{carleo2019mlinphysical}
Carleo G, Cirac I, Cranmer K, Daudet L, Schuld M, Tishby N, et~al.
\newblock Machine Learning and the Physical Sciences.
\newblock Reviews of Modern Physics. 2019;91(4):045002.
\newblock doi:{10.1103/RevModPhys.91.045002}.

\bibitem{guidotti:2018:a}
Guidotti R, Monreale A, Ruggieri S, Turini F, Giannotti F, Pedreschi D.
\newblock A {{Survey}} of {{Methods}} for {{Explaining Black Box Models}}.
\newblock ACM Computing Surveys. 2019;51(5):1--42.
\newblock doi:{10.1145/3236009}.

\bibitem{linardatos2021explainable}
Linardatos P, Papastefanopoulos V, Kotsiantis S.
\newblock Explainable {{AI}}: {{A Review}} of {{Machine Learning
  Interpretability Methods}}.
\newblock Entropy. 2021;23(1):18.
\newblock doi:{10.3390/e23010018}.

\bibitem{ribeiro2016}
Ribeiro MT, Singh S, Guestrin C.
\newblock "{{Why Should I Trust You}}?": {{Explaining}} the {{Predictions}} of
  {{Any Classifier}}.
\newblock In: Proceedings of the 22nd {{ACM SIGKDD International Conference}}
  on {{Knowledge Discovery}} and {{Data Mining}}. {ACM}; 2016. p. 1135--1144.

\bibitem{Lundberg_Lee_2017}
Lundberg SM, Lee SI.
\newblock A {{Unified Approach}} to {{Interpreting Model Predictions}}.
\newblock In: Advances in {{Neural Information Processing Systems}}. vol.~30.
  {Curran Associates, Inc.}; 2017.Available from:
  \url{https://proceedings.neurips.cc/paper/2017/file/8a20a8621978632d76c43dfd28b67767-Paper.pdf}.

\bibitem{bjorklund2023explaining}
Bj{\"o}rklund A, Henelius A, Oikarinen E, Kallonen K, Puolam{\"a}ki K.
\newblock Explaining Any Black Box Model Using Real Data.
\newblock Frontiers in Computer Science. 2023;5:1143904.
\newblock doi:{10.3389/fcomp.2023.1143904}.

\bibitem{lipton2018mythos}
Lipton ZC.
\newblock The {{Mythos}} of {{Model Interpretability}}: {{In}} Machine
  Learning, the Concept of Interpretability Is Both Important and Slippery.
\newblock Queue. 2018;16(3):31--57.
\newblock doi:{10.1145/3236386.3241340}.

\bibitem{wellawatte2023perspective}
Wellawatte GP, Gandhi HA, Seshadri A, White AD.
\newblock A {{Perspective}} on {{Explanations}} of {{Molecular Prediction
  Models}}.
\newblock Journal of Chemical Theory and Computation. 2023;19(8):2149--2160.
\newblock doi:{10.1021/acs.jctc.2c01235}.

\bibitem{hooker2021unrestricted}
Hooker G, Mentch L, Zhou S.
\newblock Unrestricted Permutation Forces Extrapolation: Variable Importance
  Requires at Least One More Model, or There Is No Free Variable Importance.
\newblock Statistics and Computing. 2021;31(6):82.
\newblock doi:{10.1007/s11222-021-10057-z}.

\bibitem{bjorklund2023slisemap}
Bj{\"o}rklund A, M{\"a}kel{\"a} J, Puolam{\"a}ki K.
\newblock {{SLISEMAP}}: Supervised Dimensionality Reduction through Local
  Explanations.
\newblock Machine Learning. 2023;112(1):1--43.
\newblock doi:{10.1007/s10994-022-06261-1}.

\bibitem{levine2020sensebert}
Levine Y, Lenz B, Dagan O, Ram O, Padnos D, Sharir O, et~al.
\newblock {{SenseBERT}}: {{Driving Some Sense}} into {{BERT}}.
\newblock In: Proceedings of the 58th {{Annual Meeting}} of the {{Association}}
  for {{Computational Linguistics}}. {Association for Computational
  Linguistics}; 2020. p. 4656--4667.

\bibitem{pearson1901liii}
Pearson K.
\newblock {{LIII}}. {{On}} Lines and Planes of Closest Fit to Systems of Points
  in Space.
\newblock The London, Edinburgh, and Dublin Philosophical Magazine and Journal
  of Science. 1901;2(11):559--572.
\newblock doi:{10.1080/14786440109462720}.

\bibitem{sorzano2014survey}
Sorzano COS, Vargas J, Montano AP. A Survey of Dimensionality Reduction
  Techniques; 2014.

\bibitem{vogelstein2021Supervised}
Vogelstein JT, Bridgeford EW, Tang M, Zheng D, Douville C, Burns R, et~al.
\newblock Supervised Dimensionality Reduction for Big Data.
\newblock Nature Communications. 2021;12(1):2872.
\newblock doi:{10.1038/s41467-021-23102-2}.

\bibitem{joia2011affine}
Joia P, Coimbra D, Cuminato JA, Paulovich FV, Nonato LG.
\newblock Local Affine Multidimensional Projection.
\newblock IEEE Transactions on Visualization and Computer Graphics.
  2011;17(12):2563--2571.
\newblock doi:{10.1109/TVCG.2011.220}.

\bibitem{leeuw2015shepard}
Leeuw JD, Mair P.
\newblock Shepard Diagram.
\newblock In: Wiley {{StatsRef}}: {{Statistics}} Reference Online. {John Wiley
  \& Sons, Ltd}; 2015. p. 1--3.

\bibitem{JMLR:v11:venna10a}
Venna J, Peltonen J, Nybo K, Aidos H, Kaski S.
\newblock Information Retrieval Perspective to Nonlinear Dimensionality
  Reduction for Data Visualization.
\newblock Journal of Machine Learning Research. 2010;11(13):451--490.

\bibitem{espadoto2021Quantitative}
Espadoto M, Martins RM, Kerren A, Hirata NST, Telea AC.
\newblock Toward a {{Quantitative Survey}} of {{Dimension Reduction
  Techniques}}.
\newblock IEEE Transactions on Visualization and Computer Graphics.
  2021;27(3):2153--2173.
\newblock doi:{10.1109/TVCG.2019.2944182}.

\bibitem{bjorklund2019}
Bj{\"o}rklund A, Henelius A, Oikarinen E, Kallonen K, Puolam{\"a}ki K.
\newblock Sparse {{Robust Regression}} for {{Explaining Classifiers}}.
\newblock In: Discovery {{Science}}. vol. 11828. Springer; 2019. p. 351--366.

\bibitem{nocedal1980updating}
Nocedal J.
\newblock Updating Quasi-{{Newton}} Matrices with Limited Storage.
\newblock Mathematics of Computation. 1980;35(151):773--782.
\newblock doi:{10.1090/S0025-5718-1980-0572855-7}.

\bibitem{bjorklund2023slisemapdemo}
Bj{\"o}rklund A, M{\"a}kel{\"a} J, Puolam{\"a}ki K.
\newblock {{SLISEMAP}}: {{Combining Supervised Dimensionality Reduction}} with
  {{Local Explanations}}.
\newblock In: Machine {{Learning}} and {{Knowledge Discovery}} in
  {{Databases}}. vol. 13718. {Springer}; 2023. p. 612--616.

\bibitem{tanaka2023chi}
Tanaka A, Tyree J, Bj{\"o}rklund A, M{\"a}kel{\"a} J, Puolam{\"a}ki K.
\newblock $\chi$iplot: {{Web-First Visualisation Platform}} for
  {{Multidimensional Data}}.
\newblock In: De~Francisci~Morales G, Perlich C, Ruchansky N, Kourtellis N,
  Baralis E, Bonchi F, editors. Machine {{Learning}} and {{Knowledge
  Discovery}} in {{Databases}}: {{Applied Data Science}} and {{Demo Track}}.
  vol. 14175. {Cham}: {Springer Nature Switzerland}; 2023. p. 335--339.

\bibitem{roberttibshiraniRegressionShrinkageSelection1996}
Tibshirani R.
\newblock Regression {{Shrinkage}} and {{Selection}} via the {{Lasso}}.
\newblock Journal of the Royal Statistical Society Series B (Methodological).
  1996;58(1):267--288.

\bibitem{besel2023GeckoQ}
Besel V. {{GeckoQ}}: {{Atomic}} Structures, Conformers and Thermodynamic
  Properties of 32k Atmospheric Molecules; 2023.

\bibitem{Nilakantan87}
Nilakantan R, Bauman N, Dixon JS, Venkataraghavan R.
\newblock Topological torsion: a new molecular descriptor for SAR applications.
  Comparison with other descriptors.
\newblock Journal of Chemical Information and Computer Sciences.
  1987;27(2):82--85.
\newblock doi:{10.1021/ci00054a008}.

\bibitem{rdkit}
Landrum G. RDKit: Open-source cheminformatics; 2006.
\newblock Available from: \url{https://www.rdkit.org}.

\bibitem{kerminen2018}
Kerminen VM, Chen X, Vakkari V, Petäjä T, Kulmala M, Bianchi F.
\newblock Atmospheric new particle formation and growth: review of field
  observations.
\newblock Environmental Research Letters. 2018;13(10):103003.
\newblock doi:{10.1088/1748-9326/aadf3c}.

\bibitem{Klamt93}
Klamt A, Schüürmann G.
\newblock COSMO: a new approach to dielectric screening in solvents with
  explicit expressions for the screening energy and its gradient.
\newblock J Chem Soc{,} Perkin Trans 2. 1993; p. 799--805.
\newblock doi:{10.1039/P29930000799}.

\bibitem{Klamt98}
Klamt A, Jonas V, Bürger T, Lohrenz JCW.
\newblock Refinement and Parametrization of COSMO-RS.
\newblock The Journal of Physical Chemistry A. 1998;102(26):5074--5085.
\newblock doi:{10.1021/jp980017s}.

\bibitem{cms2017dataset}
{CMS Collaboration}. Simulated Dataset \{\vphantom\}{{QCD}}\_{{Pt}}
  -15to3000\_{{TuneZ2star}}\_{{Flat}}\_{{8TeV}}\_pythia6\vphantom\{\} in
  \{\vphantom\}{{AODSIM}}\vphantom\{\} Format for 2012 Collision Data; 2017.

\bibitem{cms2013performance}
{CMS Collaboration}.
\newblock {Performance of quark/gluon discrimination in 8 TeV pp data}.
\newblock Geneva: CERN; 2013.
\newblock Available from: \url{https://cds.cern.ch/record/1599732}.

\bibitem{breiman2001random}
Breiman L.
\newblock Random {{Forests}}.
\newblock Machine Learning. 2001;45(1):5--32.
\newblock doi:{10.1023/A:1010933404324}.

\bibitem{blum2009}
Blum LC, Reymond JL.
\newblock 970 Million Druglike Small Molecules for Virtual Screening in the
  Chemical Universe Database {GDB-13}.
\newblock J Am Chem Soc. 2009;131:8732.

\bibitem{rupp2012}
Rupp M, Tkatchenko A, M\"uller KR, von Lilienfeld OA.
\newblock Fast and accurate modeling of molecular atomization energies with
  machine learning.
\newblock Physical Review Letters. 2012;108:058301.

\bibitem{stuke2019chemical}
Stuke A, Todorovi{\'c} M, Rupp M, Kunkel C, Ghosh K, Himanen L, et~al.
\newblock Chemical Diversity in Molecular Orbital Energy Predictions with
  Kernel Ridge Regression.
\newblock The Journal of Chemical Physics. 2019;150(20):204121.
\newblock doi:{10.1063/1.5086105}.

\bibitem{ghosh2020qm9}
Ghosh K. {MBTR\_QM9}; 2020.
\newblock Available from: \url{https://doi.org/10.5281/zenodo.4035918}.

\bibitem{Moriwaki2018}
Moriwaki H, Tian YS, Kawashita N, Takagi T.
\newblock Mordred: a molecular descriptor calculator.
\newblock Journal of Cheminformatics. 2018;10(1):4.
\newblock doi:{10.1186/s13321-018-0258-y}.

\end{thebibliography}

\newpage
\section*{Supporting information}
\beginsupplement
\section{Chosen explainable features}
\label{sec:exp_feats}
The following variables available from the GeckoQ data were utilized as \textit{Explainable features}: Number of atoms, number of carbon atoms, C=C (non-aromatic, hydroxyl (alkyl), aldehyde, ketone, carboxylic acid, ester, ether (alicyclic), nitrate, nitro, carbonylperoxynitrate, peroxide, hydroperoxide, carbonylperoxyacid. The functional groups 'C=C-C=O in non-aromatic ring', 'aromatic hydroxyl', 'nitroester' were also available. Still, we decided to drop them because fewer than 2 \% of the data had an entry for these properties.

\section{Incidence of functional groups by cluster in GeckoQ}
\label{sec:incidence}
\begin{figure}[!b]
    \centering
    \includegraphics[width=\textwidth]{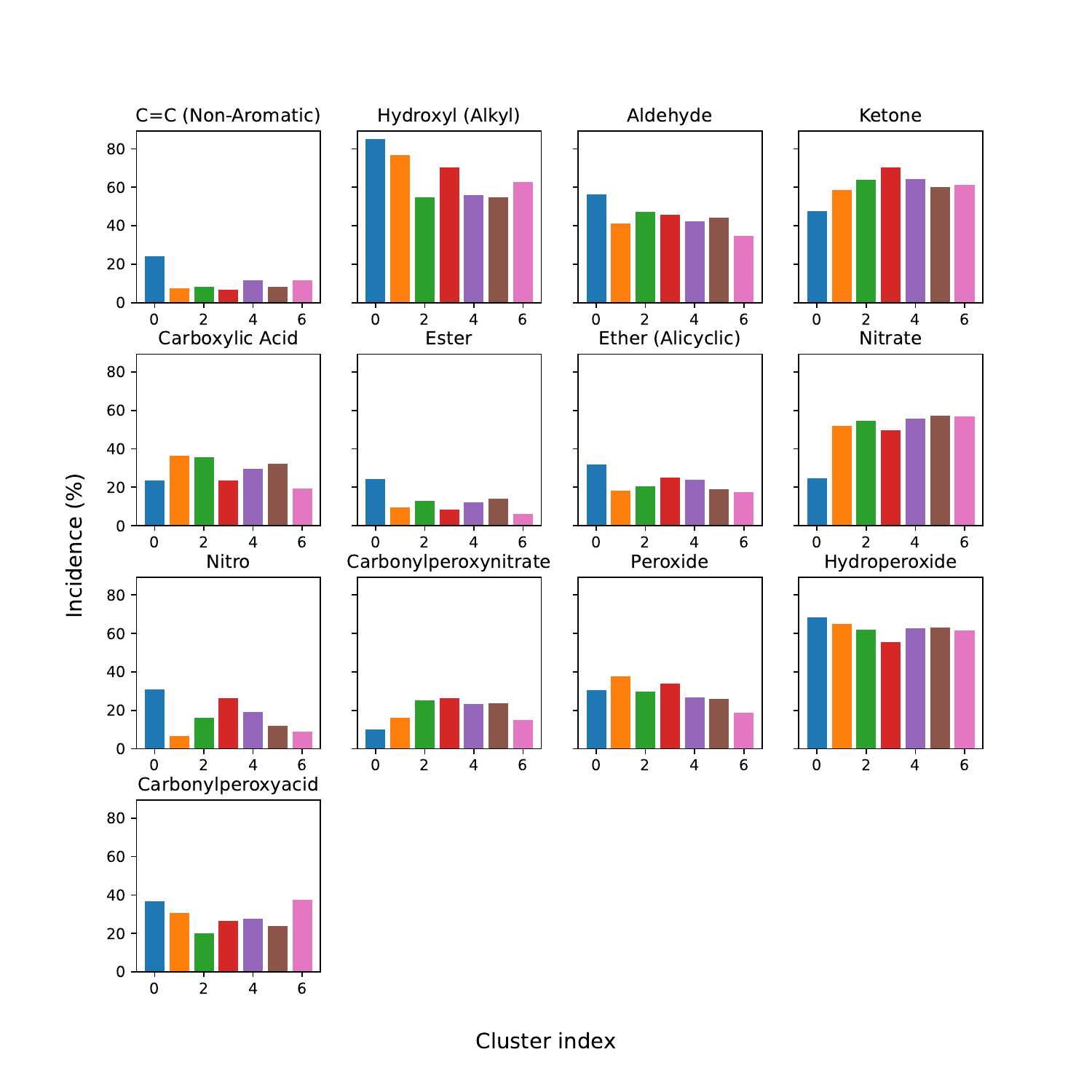}
    \caption{Functional groups by cluster.}
    \label{fig:incidence_by_cluster}
\end{figure}

See \autoref{fig:incidence_by_cluster} for incidence of functional groups for each of the clusters described in Section 4.

\section{Explanation metrics}
\label{sec:explanation_metrics}
\textit{Local loss.} The local loss (called ``fidelity'' in \cite{bjorklund2023slisemap}) measures how well the local models learned by {\sc slisemap} predict the target.
Mathematically, local loss is expressed as
\begin{equation*}
    \frac{1}{n} \sum_{i=1}^{n} l(f_i(\mathbf{x}_i, \mathbf{y}_i),
\end{equation*}
where $f_i$ are the local models and $l$ is the loss function as defined in \autoref{sec:slisemap}.
We are not only interested on the performance of the individual local model-target pairs but also how well the local models can predict the neighbouring points in the embedding.
To find the neighbours, we use the $k$ nearest neighbours, rendering
\begin{equation*}
    \frac{1}{n} \sum_{i=1}^{n} \frac{1}{k} \sum_{j \in \textrm{k-NN}(i)} l(f_i(\mathbf{x}_j, \mathbf{y}_j).
\end{equation*}
In this paper, we chose to use $k = \lfloor 0.1N\rfloor$ where $N$ is the number of data items.
A smaller value of local loss is better.

\textit{Coverage.}
Coverage \cite{guidotti:2018:a} measures how well local models generalise to other data points.
It can be calculated by counting the number of data items that have a local loss less than a set threshold $l_0$:
\begin{equation*}
    \frac{1}{n} \sum_{i=1}^{n} \frac{1}{n} \sum_{j=1}^{n} |\{j : l(g_i(\mathbf{x}_j), \mathbf{y}_j) < l_0 \} |.
\end{equation*}
We are more interested, however, about the local coverage of the models.
Thus, we limit the coverage testing to the $k$ nearest neighbours as above, and with the same value of $k$.
We get
\begin{equation*}
    \frac{1}{n} \sum_{i=1}^{n} \frac{1}{k} \sum_{j \in \textrm{k-NN}(i)} |\{j : l(g_i(\mathbf{x}_j), \mathbf{y}_j) < l_0 \} |.
\end{equation*}
In this paper, we have chosen the loss threshold to be the $0.3$ quantile of the losses of a global linear model (similar to \cite{bjorklund2023slisemap}).
A higher coverage indicates better performance.

\end{document}